\documentclass[letterpaper, 10 pt, conference]{ieeeconf}
\IEEEoverridecommandlockouts 
\usepackage{amssymb}
\usepackage[colorlinks,linkcolor=blue]{hyperref}
\usepackage{amsmath}
\usepackage{graphicx}
\usepackage{caption}
\usepackage{subcaption}
\graphicspath{ {images/} }
\usepackage{float}

\usepackage{amsthm}
\newcommand{\norm}[1]{\left\lVert#1\right\rVert}
\newtheorem{theorem}{Theorem}

\newtheorem{definition}[theorem]{Definition}

\usepackage{ upgreek }
\usepackage{algorithm}
\usepackage{algorithmicx}
\usepackage{algpseudocode}
\usepackage{url}
\usepackage{soul}
\usepackage{lipsum}
\usepackage{mathrsfs}

\newcommand{\R}{\mathbb{R}}

\newcommand{\uu}{\mathbf{u}}
\newcommand{\s}{\mathbf{s}}
\newcommand{\p}{\mathbf{p}}
\newcommand{\sig}{\mathbf{\sigma}}

\newcommand{\Ss}{\mathcal{S}}

\title{\LARGE \bf
A Modeled Approach for Online Adversarial Test of \\ Operational Vehicle Safety (extended version)
}

\author{
Linda Capito$^{*1}$, Bowen Weng$^{*1}$, \\ Umit Ozguner$^1$, \textit{IEEE Life Fellow}, Keith Redmill$^1$, \textit{IEEE Senior Member and SIAM Member}
\thanks{$^*$These authors contributed equally.}
\thanks{$^1$The authors are with the Department of Electrical and Computer Engineering at Ohio State University, OH, USA.}
\thanks{Corresponding author: Keith Redmill, redmill.1@osu.edu}
}

\begin{document}

\maketitle
\thispagestyle{empty}
\pagestyle{empty}

\begin{abstract}
    The scenario-based testing of operational vehicle safety presents a set of principal other vehicle (POV) trajectories that seek to force the subject vehicle (SV) into a certain safety-critical situation. Current scenarios are mostly (i) statistics-driven: inspired by human driver crash data, (ii) deterministic: POV trajectories are pre-determined and are independent of SV responses, and (iii) overly simplified: defined over a finite set of actions performed at the abstracted motion planning level. Such scenario-based testing (i) lacks severity guarantees, (ii) has predefined maneuvers making it easy for an SV with intelligent driving policies to game the test, and (iii) is inefficient in producing safety-critical instances with limited and expensive testing effort. We propose a model-driven online feedback control policy for multiple POVs which propagates efficient adversarial trajectories while respecting traffic rules and other concerns formulated as an admissible state-action space. The approach is formulated in an anchor-template hierarchy structure, with the template model planning inducing a theoretical SV capturing guarantee under standard assumptions. The planned adversarial trajectory is then tracked by a lower-level controller applied to the full-system or the anchor model. The effectiveness of the methodology is illustrated through various simulated examples with the SV controlled by either parameterized self-driving policies or human drivers.
\end{abstract}


\section{Introduction}
\begin{figure}[t!]
\vspace{3mm}
\centering
\begin{subfigure}{.49\linewidth}
  \centering
  \includegraphics[width=1\textwidth]{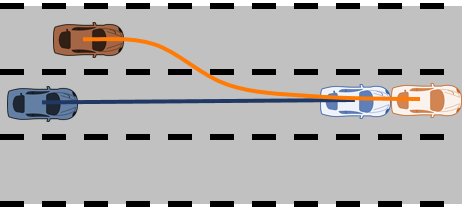}
  \caption{A standard Lead Vehicle Lane Change and Braking (LVLCB) test scenario.}
  \label{fig:lvlcb}
\end{subfigure}
\begin{subfigure}{.49\linewidth}
  \centering
  \includegraphics[width=1\textwidth]{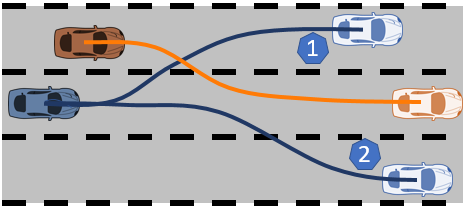}
  \caption{Two example trajectories of SV maneuvers that would "hack" the standard LVLCB test.}
  \label{fig:hack_lvlcb}
\end{subfigure}
\begin{subfigure}{.49\linewidth}
  \centering
  \includegraphics[width=1\textwidth]{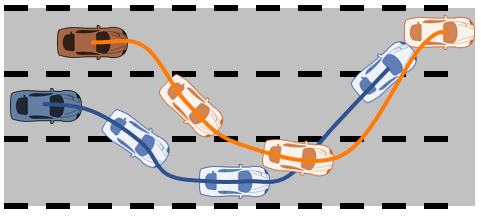}
  \caption{An adversarial POV will restrict the scenario propagation in a safety-critical status and force the SV into collisions if it is not responding appropriately.}
  \label{fig:adv_lvlcb_crash}
\end{subfigure}
\begin{subfigure}{.49\linewidth}
  \centering
  \includegraphics[width=1\textwidth]{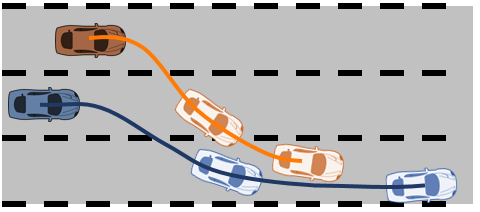}
  \caption{One option for the SV to avoid collisions is to perform extreme evasive maneuvers that are beyond the expectation of the formulation.}
  \label{fig:adv_lvlcb_nocrash}
\end{subfigure}
\caption{Various maneuvers inspired by the standard LVLCB scenario~\cite{nhtsa2019tja}: inspired by a commonly observed cause of rear-end accident from human driver crash data, the POV performs a single lane-change to get in front of the SV, followed by an immediate braking maneuver.}
\vspace{-5mm}
\label{fig:lvlcb-examples}
\end{figure}

A scenario-based evaluation method "attacks" the test subject through enforcing it into a safety-critical situation and observes the subject response and testing outcomes. For operational vehicle safety evaluation, especially concerning Advanced Driver Assistance System (ADAS)~\cite{paul2016advanced} and Automated Driving Systems (ADS)~\cite{thorn2018framework}, the design and execution of test scenarios have raised wide attention from the research community~\cite{feng2020testing}, automotive industry, and government sectors for policy making~\cite{thorn2018framework, nhtsa2019tja, nhtsa2010cib}. Typically, one first extracts initialization conditions and vehicle maneuvers inspired by the statistical observation of human driver crash data. The scenario is then specified with a given initialization state for all vehicles and a set of pre-determined principal other vehicle (POV) trajectories. Note that the POV trajectory is generated through a series of high-level planning commands and is often independent of the subject vehicle (SV) response. For example, one can refer to Fig.~\ref{fig:lvlcb} that shows a Lead Vehicle Lane Change and Braking (LVLCB) scenario~\cite{nhtsa2019tja} for ADAS safety evaluation.


The problems of the aforementioned scenario designs are threefold. First, the current scenarios lack severity guarantees. The assumption that a scenario extracted as described above and brought to a test track will generate similar SV responses, similar safety-critical situations, and similar outcomes ignores the significant differences in sensing, control algorithms, control authority and implementation on the different SVs being tested. Second, with pre-determined POV trajectories defined over a simplified action space it is easy for the SV to game the test, especially given that the SV is operating with intelligent decision making both longitudinally and laterally (e.g., Fig.~\ref{fig:hack_lvlcb}). Finally, the current approach does not create sufficient safety-critical instances within the limited and expensive real-world testing effort. While virtual simulations have provided the opportunity to enhance autonomous vehicle testing efficiency, the sim-to-real gap remains a challenge in general. Furthermore, the simplified action space is incapable of creating agile, dynamic maneuvers, which confines the scenario propagation to relatively simple configurations. This also limits the testing efficiency in both real-world tests and simulations.

Inspired by the problems mentioned above, the adversarial adaptive testing scheme has become a natural alternative. Existing approaches mostly focus on learning inspired methods~\cite{klischat2019generating, chen2020adversarial, ding2020learning}, which typically require the algorithm to interact with a fixed SV driving algorithm for many iterations. These are sampling inefficient and would only be applicable in virtual simulations. Some also seek for guided sampling from an offline-built scenario library~\cite{feng2020testing}. While this approach resolves the traditional scenario-based testing approach's problem by making it more difficult for the SV to game the test, the other two problems mentioned above remain valid.


To address the aforementioned problems of both the traditional scenario-based testing methods and the adaptive adversarial testing approaches, the proposed methodology is inspired by investigating the following two questions.
\begin{list}{$\bullet$}{\topsep=0.ex \leftmargin=0.1in \rightmargin=0.in \itemsep =0.02in}
\item {\em Q1: If we can predict the SV motion trajectory with sufficient accuracy, can we design the POV motion to reach a safety-critical scenario?}
\end{list}
In general, if the infinite-time future planning of the SV is perfectly known and the POV is sufficiently capable in terms of its  admissible space for control, the POV is  guaranteed to at least asymptotically capture the SV, i.e. create conditions that lead to inevitable collisions between the POV and the SV. This is achieved by taking the known SV motion trajectory as a motion tracking reference for the POV, which naturally leads to a minimization problem closely related to the Model Predictive Control (MPC) formulation~\cite{camacho2013model}.

\begin{list}{$\bullet$}{\topsep=0.ex \leftmargin=0.1in \rightmargin=0.in \itemsep =0.02in}
\item {\em Q2: If the SV motion trajectory is unknown or difficult to predict, can we design the POV motion to arrive at a safety-critical scenario?}
\end{list}
The reachability analysis from differential games~\cite{ho1965differential} and optimal control has answered the above question analytically. In general, the capture of the SV cannot be guaranteed if the SV's policy is unknown. However, suppose the SV and POV are sufficiently close such that the relative state belongs to the maximal backward reachable set (see equation (6) in~\cite{mitchell2007comparing}). In that case, there exists a feedback control policy for the POV such that for all possible motion trajectories of the SV, the policy renders a motion trajectory that guarantees the capturing of the SV. By assigning such a worst-case policy to the POV, one can ensure the generation of a safety-critical scenario.

In practice, one should also factorize the admissible action constraints, the operable domain allowed by traffic rules, and other concerns to propagate a "reasonable" dangerous scenario. Combining the solutions for the above questions, we propose an online feedback control policy for POVs that allows rendering adversarial paths without breaking traffic rules or the conventions of traffic accident fault or liability. The contribution of this work is threefold:

\paragraph{\textbf{A Modeled Approach}} The proposed framework is model-driven and induces theoretical guarantees for a certain safety-critical level of the derived scenario under standard assumptions. 
    
\paragraph{\textbf{Adaptive Scenarios}} The derived motion trajectory of each POV varies adaptively based on the SV response throughout the propagation of a testing scenario. This creates more challenging testing instances with the same amount of testing effort as many of the current ADS testing scenario designs~\cite{nhtsa2010cib, nhtsa2019tja} where the POV actions are independent of the SV behavior. The proposed adaptive framework acts on the POV control level, which also creates agile motion trajectories that make it more difficult for the SV to game the scenario.
    
\paragraph{\textbf{Online Execution}} The proposed method formulates an anchor-template hierarchy control structure~\cite{wensing2017template}, with the simplified template planning naturally leading to a series of quadratic programming problems with computationally tractable solutions. This is, to the best of our knowledge, the first introduction of an online path planning algorithm for POVs which constructs adversarial testing scenarios. 

Furthermore, we extensively evaluate the proposed framework's performance in a multi-lane straight-segment environment with parameterized SV driving policies and human drivers of various levels of aggressiveness. 
We observe that the adaptive framework enforces the SV into various safety-critical situations. One particularly interesting observation is that the SV typically remains collision-free if (i) the driving policy is conservatively safe, or (ii) the SV is willing to take extreme evasive maneuvers that are beyond the expectation of the modeled formulation (see Fig.~\ref{fig:adv_lvlcb_nocrash}).

The remainder of this paper is organized as follows. Section \ref{sec:prelim} introduces the background setting of this work, including the problem formulation. Section \ref{sec:method} presents details on the proposed method. Section \ref{sec:res} explains the experimental settings and the results analysis and finally, section \ref{sec:concl} presents conclusions and future work.

\section{Preliminaries} \label{sec:prelim}
We present the problem formulation of operational vehicle safety evaluation through the scenario-based test. We also revisit the basics of the anchor-template hierarchy control framework.

\subsection{Problem formulation}
Consider a heterogeneous multi-vehicle system of one subject vehicle (SV) and $k$ principal other vehicles (POVs). Each vehicle's motion is subject to a certain ordinary differential equation in general. Without loss of generality, for the $i$-th individual agent ($i \in \{0, \ldots, k\}$), 
\begin{equation} \label{eq:sysdyn}
    \dot{\s_i}=f_i(\s_i,\uu_i,t), \s_i \in \Ss_i \subset \R^n , \uu_i \in \mathcal{U}_i \subset \R^m.
\end{equation}
For the remainder of this section, the subscript $i$ is omitted for general discussions of vehicle motion. We consider that the map $f:S\times U \to S$ is uniformly continuous, bounded and Lipschitz continuous in $\s$ for fixed control $\uu$. Hence, there exists a unique trajectory solving \eqref{eq:sysdyn} for a given measurable control action $\uu \in \mathcal{U}$.

Let a \emph{snapshot} $\mathbf{\sigma} \in \Ss^{k+1}$ be the combined states of all vehicles at a time instance~\cite{weng2020model}. For example, the snapshot or state space of a test that involves two vehicles on straight segments would consist of the initial velocity of both vehicles and the relative position between them.

We then have the following definition of a \emph{testing scenario} for $k+1$ participants:
\begin{definition}
A scenario is an automaton denoted by a tuple of $\langle \Sigma, \Pi, f, \sig_0, \Lambda \rangle$, where
\begin{itemize}
    \item $\Sigma = \Ss^{k+1}$ is the set of snapshots, with each snapshot containing one SV and $k$ POVs as specified above.
    \item $\Pi$ is the "alphabet" of the automaton, in particular, the control action space of all vehicular participants in the scenario, i.e., $\Pi = \mathcal{U}^{k+1}$.
    \item $f$ is the motion transition function as defined in~\eqref{eq:sysdyn}.
    \item $\sig_0$ specifies the initial condition, i.e. the arbitrary given snapshot.
    \item $\Lambda$ is the subset of $\Sigma$ deemed acceptable.
\end{itemize}
\end{definition}
We define the acceptable set of snapshots $\Lambda$, containing all the snapshots that comply with the safety requirement, as

\begin{equation}\label{eq:safe-set}
    \Lambda = \{ \mathbf{\sigma} \in \Sigma \mid \norm{\p_0 - \p_j}_2 \geq c, \forall j \in \{1, \ldots, k\} \},
\end{equation}
where $\p \in \R^2$ denotes the position states and $c$ is referred to as the \emph{capture diameter}. Note that only the safety of the SV is of interest here. Correspondingly, the unsafe set of states is 
\vspace{-2.5mm}
\begin{equation} \label{eq:unsafe-set}
    \Omega = \Sigma \setminus \Lambda.
\end{equation}

Given a string of control inputs, we designate the \emph{run} of a scenario be a chronological sequence of snapshots $\mathscr{R}=\{\sig_i\}_{i=0,..,T}$ from initial to final time $T$.

Most of the elements that define a scenario are given, with only the initial condition $\sig_0$ and the automaton alphabet $\Pi$ subject to various possibilities of configurations. Given that SV control is one of the subjects of the test, only the POV control actions are of interest for scenario designs. Intuitively, one seeks to derive a feedback control policy 

\begin{equation}\label{eq:pov_pi}
    \pi: \Sigma \times \mathcal{U}^k \rightarrow \Sigma    
\end{equation}
for the group of POVs such that the SV is forced into a sufficiently dangerous situation. Although the POV control is specified through a general class of functions that could be state-dependent and time-variant, it is worth emphasizing that the typical implementation of the POV control in practice remains fairly simple. POV action trajectories are mostly independent of state propagation, as shown in Fig.~\ref {fig:lvlcb}. 

In general, the multi-vehicle system is subject to highly nonlinear motion dynamics and complex state-action constraints. From the model point of view, it remains a challenge to derive agile motion planning algorithms for multiple POVs to cooperate in forcing the SV into dangerous situations. In this paper, we consider an anchor-template framework, as introduced in the following section.

\subsection{Anchor-template framework for vehicle motion planning}
When a full system model or anchor model is too complex to permit the design an appropriate controller, it is possible to use a simplification or template model that still captures the anchor model's essential characteristics and properties. Such a framework has been widely adopted in bipedal locomotion~\cite{wieber2006trajectory} and vehicle control~\cite{rajamani2011vehicle}.

A typical template model captures the motion features of the anchor model with lower dimensional state space and simplified dynamics. In this paper, the complex anchor model state is replaced with the simpler template state of discrete-time locally linearized motion equation formulated as
\begin{equation}
    \s(t+\Delta) = \mathbf{A}\s(t) + \mathbf{B}\uu(t),    
\end{equation}
with time-step $\Delta$ and system matrices
\begin{equation} \label{eq:linear}
    \mathbf{A} =  \begin{bmatrix} 1 & 0 & \Delta & 0 \\ 0 & 1 & 0 & \tilde{v} \Delta \\ 0 & 0 & 1 & 0  \\ 0 & 0 & 0 & 1 \end{bmatrix} , \quad \mathbf{B} =   \begin{bmatrix}  0 & 0 \\ 0 & 0 \\ \Delta & 0 \\ 0 & \frac{\Delta}{\tilde{v}} \end{bmatrix}. 
\end{equation}
The state vector $\s \in \R^4$ for each vehicle consists of the Cartesian coordinates of its position $\mathbf{p}=[x,y]$, speed $v$ and heading angle $\phi$. For the linearization, a constant speed $\tilde{v}$ is also used. The control action includes longitudinal and lateral accelerations $\uu=\begin{bmatrix} a_x,a_y \end{bmatrix}^T$. 

The model above is adapted from~\cite{junietz2018criticality} and \cite{weng2020model} assuming small local course angles (i.e. the angular difference between the combined velocity vector and the vehicle's running direction) and small changes in speed for local linearization. Theoretically, the control input $\uu$ is subject to the Kamm's circle~\cite{erlien2015shared} induced bounds. The state constraints are typically determined by speed limits, road topology, and other concerns regarding the admissible operable domain of vehicles. In this paper, we approximate the state-action constraints with a linear architecture in the form of $\mathbf{G}_u\uu \leq \mathbf{h}_u$ and $\mathbf{G}_s\s \leq \mathbf{h}_s$. 

\section{Method} \label{sec:method}
\begin{figure}[t!]
\centering
\vspace{3mm}
\includegraphics[width=0.9\linewidth]{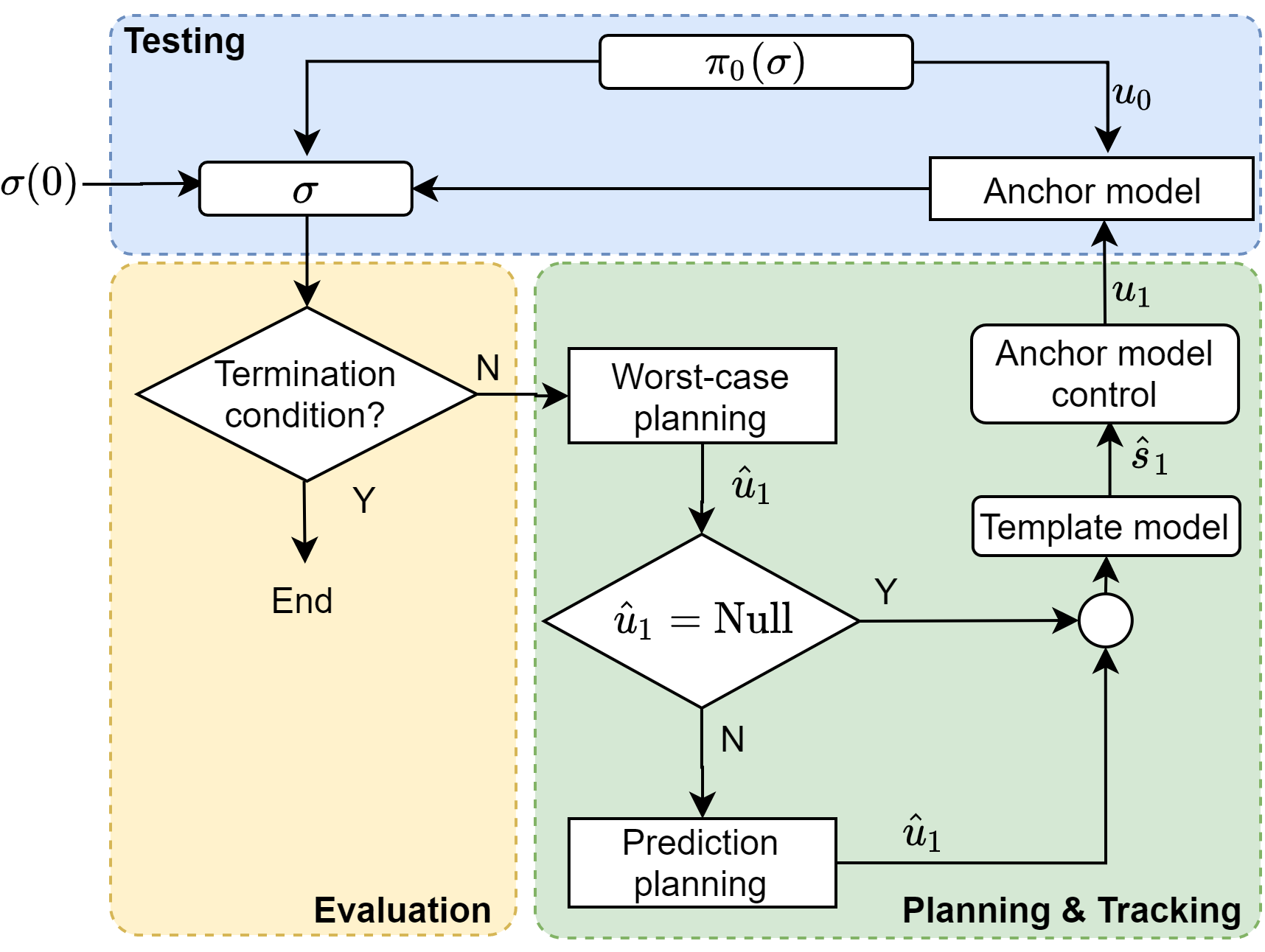}
\caption{Proposed method flowchart under the pairwise interactive setting with one POV and one SV.}
\label{fig:flowchart}
\vspace{-2.5mm}
\end{figure}
Fig.~\ref{fig:flowchart} presents a general overview of the proposed framework. The adversarial scenario propagation starts with an initial snapshot of the system $\sigma_0$. The test ends if a certain termination condition is satisfied (e.g., encountering a collision); otherwise, one executes the template-based planning stage. For adversarial motion planning, one first seeks to create safety-critical scenarios assuming the SV policy is unknown. If the POV-to-SV distances are not sufficiently small, such a worst-case planning algorithm may not be applicable. One then moves on to the predictive planning stage, where the POV's adversarial policy is derived based on a certain assumed predicted behavior of the SV. The planned reference trajectory is then tracked by a lower-level controller such as MPC.

For the remainder of this section, we introduce the worst-case planning and predictive planning in a pairwise manner with one POV and one SV. Without loss of generality, let the subscript $1$ denote the adversarial POV and $0$ the SV. Finally, we will extend the idea to multiple cooperative POVs. 

\subsection{Worst-case planning}
Intuitively, the worst-case planning stage seeks to derive the POV adversarial trajectory without knowing the SV intentions. By the definition of unsafe snapshots $\Omega$ as specified in~\eqref{eq:unsafe-set}, starting from a safe initial condition $\mathbf{\sigma}_0 \in \Sigma$ at time $0$, if the pair of control strategies $( \pi_0(\mathbf{\sigma}), \pi_1(\mathbf{\sigma}))$ renders the trajectory converging to $\Omega$ at some finite time, we denote such a capture time as $T(\pi_0(\mathbf{\sigma}), \pi_1(\mathbf{\sigma}), \mathbf{\sigma}_0)$. The pairwise SV and POV form a zero-sum game, wherein the SV seeks to maximize $T(\cdot)$ and the POV seeks to minimize $T(\cdot)$. Correspondingly, we have the minimax optimal feedback strategy $\pi_0^*(\mathbf{\sigma})$ and $\pi_1^*(\mathbf{\sigma})$ satisfying the saddle condition as
\begin{equation} \label{eq:saddle-minimax}
        T(\pi_0(\mathbf{\sigma}), \pi_1^*(\mathbf{\sigma}), \mathbf{\sigma}_0) \leq T^* \leq T(\pi_0^*(\mathbf{\sigma}), \pi_1(\mathbf{\sigma}), \mathbf{\sigma}_0),
\end{equation}
with $T^* = T(\pi_0^*(\mathbf{\sigma}), \pi_1^*(\mathbf{\sigma}), \mathbf{\sigma}_0)$ referred to as the \emph{minimal capture time}. Theoretically, $T^*$ exists if and only if the initialization condition $\mathbf{\sigma}_0$ is inside the maximal backward reachable set~\cite{mitchell2007comparing}, i.e., there exists a POV control strategy such that regardless of the SV responses, one can ensure the capturing of the SV. This is guaranteed when the POV has moved to the reachable area of the SV, in this case, the inevitable collision zone. In practice, for a general nonlinear system, $T^*$ can be obtained through a discrete approximation of the Hamilton-Jacobian-Bellman partially differential equation (HJB-PDE)~\cite{Bokanowski2010}. In this paper, we seek to derive the $T^*$, if applicable, within a local look-ahead time horizon up to $\bar{T}$ seconds. With the simplified template dynamics, 
one can approximate $T^*$ by iteratively solving a series of minimax quadratic programming problems as discussed in~\cite{weng2020model}. One can also derive the function of mapping a snapshot to the corresponding $T^*$ offline. Given the minimal capture time $T^*$, we can then formulate the optimal feedback policy for the POV as
\begin{subequations}
    \begin{align}
        \hat{\uu}_1^* = & \underset{\hat{\uu}_1}{\text{argmin}} \underset{\hat{\uu}_0}{\text{max}} \norm{\p_0(T^*) - \p_1(T^*)}^2 \label{eq:minmax-opt}\\
        \text{s.t. } & \s_i(t+\Delta) = \mathbf{A} \s_i(t) + \mathbf{B} \uu_i(t), \forall i\in\{0, 1\}, \label{eq:worst-dyn}\\
        & \mathbf{G}_s\s_i(t) \leq \mathbf{h}_s, \forall t \in \{0, \ldots, T^*-\Delta\}, \label{eq:worst-state-c} \\
        & \mathbf{G}_u\uu_i(t) \leq \mathbf{h}_u, \forall t \in \{0, \ldots, T^*-\Delta\}, \label{eq:worst-action-c} \\
        & \mathbf{\sigma}(0) = \mathbf{\sigma}_0. \label{eq:worst-init}
    \end{align}
\end{subequations}
In practice, the above optimization is solved at each planning stage presenting the current snapshot as the initialization condition. The instantaneous reference motion trajectory $\hat{\s}_1$ is then obtained by propagating the template-model motion for $T^*$ seconds with the sequence of optimal actions $\hat{\uu}_1^*$. 

Note that the capturing guarantee at the template-model level is subject to some assumed state constraints~\eqref{eq:worst-state-c} and action constraints~\eqref{eq:worst-action-c} of the SV. That is, the capturing will undoubtedly fail if the SV is willing to perform extreme evasive maneuvers that are beyond its assumed capability. While such an outcome may be deemed aggressive in terms of passenger comfort and vehicle dynamic stability, it is still technically a safe choice of action from the operational safety perspective. Furthermore, consider that the definition of safety is induced by the $l$2-norm distance between vehicles as indicated by~\eqref{eq:safe-set}, which is not necessarily equivalent to vehicle-to-vehicle collisions in real-world driving. This leads to a trade-off phenomenon when determining the capture diameter $c$.  A larger choice of $c$ results in a larger maximal BRS and hence increases the likelihood of deriving a minimal capture time $T^*$.
However, this also makes it difficult for a real vehicle-to-vehicle collision to occur, leading to snapshots that are not sufficiently safety-critical. A more detailed comparison regarding this trade-off will be presented in Section~\ref{sec:res} between Fig.~\ref{fig:cib_crash_smallc} and Fig.~\ref{fig:cib_nocrash_bigc}. 

\subsection{Predictive planning} \label{sec:pred}
If the worst-case planning is not applicable (i.e., one cannot find a qualified minimal capture time $T^* \leq \bar{T}$), this implies the current snapshot is not severe enough to enable a guaranteed capturing. We propose to consider an alternative based on the assumed predictive motion of the SV.

Historically, vehicle motion prediction has been studied extensively~\cite{houenou2013vehicle, koschi2017spot}.
The various self-driving algorithms that are model-based~\cite{petrovskaya2008model,capito2020optical, park2013game} and learning-inspired~\cite{bojarski2016end,yurtsever2020integrating} are also applicable to serve as the SV motion predictor. In this paper, we predict the SV motion based on the steady-state assumption. The traditional steady-state assumption assumes fixed velocity and heading (i.e. control $\uu=\mathbf{0}$). It is also the fundamental assumption for various vehicle safety related methodologies including Time-to-Collision~\cite{lee1976theory} for safety analysis and human driver behavior characterization such as gap acceptance~\cite{yang2019examining} and lead-vehicle following distance~\cite{swaroop2001review}. In most of the examples presented in the Section~\ref{sec:res}, we adopt a modified steady-state assumption which assumes that the SV is maintaining the current control action up to the assigned time horizon. Let such a predictive policy for the SV be $\hat{\pi}_0$, one can then take the predicted motion of the SV as the tracking reference for the POV. With the running cost and termination cost determined by $\mathbf{Q}_r \in \R^{4\times 4}$ and $\mathbf{Q}_f \in \R^{4\times 4}$ respectively, and considering $\tilde{\s}(t)=\s_1(t)-\s_0(t)$, we have the following optimization problem for the predictive planning:
\begin{subequations}
    \begin{align}
        \hat{\uu}_1^* = & \underset{\hat{\uu}_1}{\text{argmin}} \sum_{t=0}^{\bar{T}-\Delta} ( \tilde{\s}(t)^T\mathbf{Q}_r\tilde{\s}(t) ) +  \tilde{\s}(\bar{T})^T\mathbf{Q}_f\tilde{\s}(\bar{T}) \label{eq:min-opt}\\
        \text{s.t. } & \s_1(t+\Delta) = \mathbf{A} \s_1(t) + \mathbf{B} \uu_1(t), \label{eq:pred-dyn}\\
        & \s_0(t+\Delta) = \mathbf{A} \s_0(t) + \mathbf{B} \hat{\pi}_0(\s_0(t)), \label{eq:pred-svdyn}\\
        & \mathbf{G}_s\s_i(t) \leq \mathbf{h}_s, \forall t \in \{0, \ldots, T^*-\Delta\}, \label{eq:pred-state-c} \\
        & \mathbf{G}_u\uu_i(t) \leq \mathbf{h}_u, \forall t \in \{0, \ldots, T^*-\Delta\}, \label{eq:pred-action-c} \\
        & \mathbf{\sigma}(0) = \mathbf{\sigma}_0. \label{eq:pred-init}
    \end{align}
\end{subequations}
The reference trajectory for the POV is generated by a  similar procedure as that shown in the worst-case planning section. 

Finally, an MPC is used as the navigation controller that allows the POV to follow the reference path obtained from the planning stage. It is worth emphasizing that although both MPC for vehicle motion tracking and the template-model based planning rely on a certain linearized motion equation, the two formulations are not necessarily the same. In our case, the template-model adopts the formulation in~\eqref{eq:linear} with the control action specified as accelerations, which are easier to identify through perceivable states with quantifiable constraints. On the other hand, the MPC adopts the formulation from~\cite{rajamani2011vehicle} with the control action defined as acceleration $a$ and steering wheel angle $\omega$, which is easier to implement for direct vehicle control tasks. The various state-action constraints deployed in the planning stage as specified in~\eqref{eq:minmax-opt} and~\eqref{eq:min-opt} are also included in the MPC formulation. Therefore, even if mild discrepancies may occur between the template-model used for planning and the anchor-model adopted for motion tracking, the constraints remain valid.

\vspace{-2.2mm}
\subsection{Multi-POVs}
With the adversarial pairwise interactive scenario presented above, we are now ready to extend the framework to work with multiple POVs. In practice, it is overly aggressive to assume that all POVs present in the snapshot will cooperatively attack the SV. The absolute cooperative POV planning also poses an extra challenge to the scenario design given none of the POVs are supposed to crash into each other during the scenario propagation. On the other hand, some existing work~\cite{weng2020model} considers a partially non-cooperative assumption allowing at most one adversarial POV with the rest of the POVs complying with the SV for collision avoidance. Although this assumption simplifies the analysis significantly and enables a pairwise study between the SV and each POV, the partial non-cooperativeness remains conservative. 

In this work, we propose a multi-POV scenario generation scheme which allows multiple POVs propagating cooperative safety-critical scenarios in a controlled manner. This is done by assigning each POV a dedicated set of state-action constraints which ensures the non-interactive motion between POVs. Section~\ref{sec:res} will present more detailed examples for multi-POV adversarial scenarios propagated with the proposed method.
\vspace{-1.8mm}
\section{Experimental results} \label{sec:res}

We illustrate the performance of the proposed approach with two types of simulations. We first demonstrate the adversarial testing scheme against a series of parameterized self-driving policies in a customized  simulation environment with the nonholonomic bicycle model of vehicle dynamics.  We then expand the test to human-driven SVs performed in the CARLA simulator~\cite{dosovitskiy2017carla}. Throughout all simulations, we have $\bar{T}=2$ for the planning and MPC horizon limit. We also assume an identical admissible action space for all POVs and the SV with $a_x^{\text{max}}=0.67, a_x^{\text{min}}=-1.7, a_y^{\text{max}}=1, a_y^{\text{min}}=-1$. Considering the straight-segment environment we are studying, the cost matrices in~\eqref{eq:min-opt} are defined to emphasize the lateral tracking accuracy satisfying $\mathbf{Q}_r= \mathbf{Q}_f = \text{diag}(1, 100, 0.1, 0.1)$.

\subsection{Testing parameterized self-driving policies} \label{sec:sim_policies}
The parameterized self-driving policy is a combination of the Intelligent Driving Model (IDM) and a set of customized lane-change heuristics. The IDM formulation is adapted from~\cite{liebner2012driver}. The lane change heuristics consist of two stages, decision making and lane-change execution. The lane-change decision is determined by the current SV velocity and the lead-lag gap acceptance with parameters adopted from the naturalistic behavioral study of~\cite{yang2019examining}. The lane-change execution is divided into two parts. The longitudinal velocity is adapted from the IDM model. The lateral motion is controlled by a parameterized PD controller subject to yaw-rate constraints. We first present a set of pairwise interactive scenarios with one POV and one SV. We then extend to the multi-POV settings.

\subsubsection{Pairwise Interactions}
\begin{figure}[b!]
\centering
\includegraphics[width=0.8\linewidth]{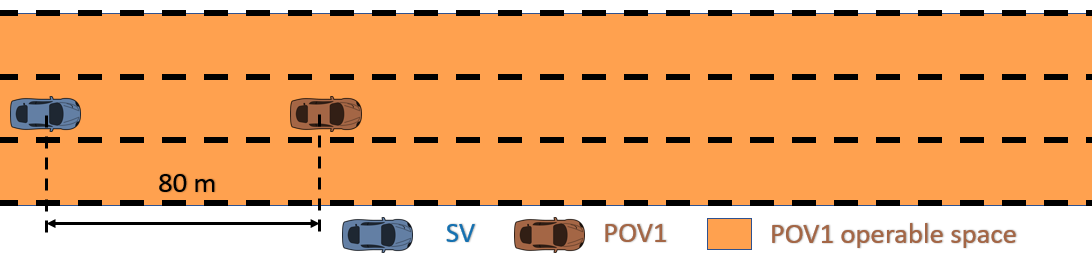}
\caption{The standard Crash Imminent Brake (CIB) system testing procedure~\cite{nhtsa2010cib}: the lead-POV brakes and expects the SV to also brake for collision avoidance: both vehicles start within the same lane with an initial velocity of $18 m/s$.}
\label{fig:cib_concept}
\end{figure}
\begin{figure}[h!]
\centering
\vspace{3mm}
\begin{subfigure}{.4\textwidth}
  \centering
  \includegraphics[width=1\textwidth]{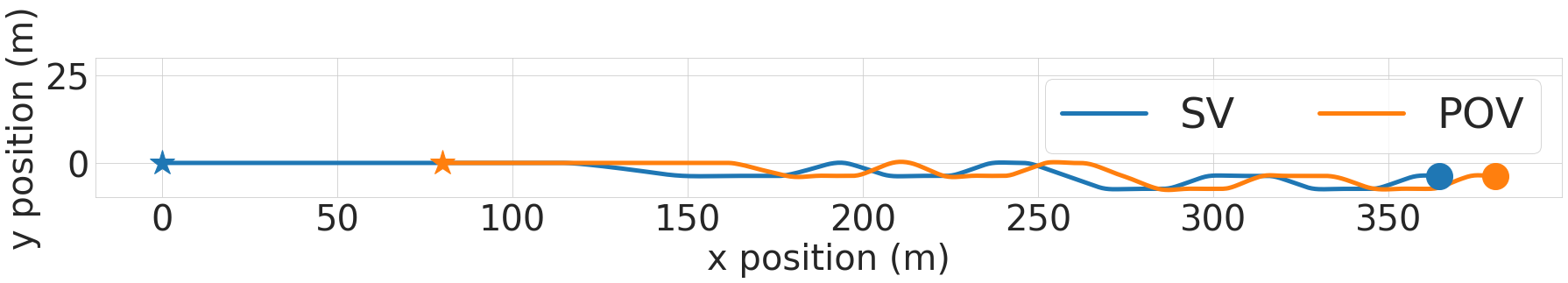}
  \caption{Vehicle position trajectories.}
  \label{fig:cib_annoying1}
\end{subfigure}
\begin{subfigure}{.4\textwidth}
  \centering
  \includegraphics[width=1\textwidth]{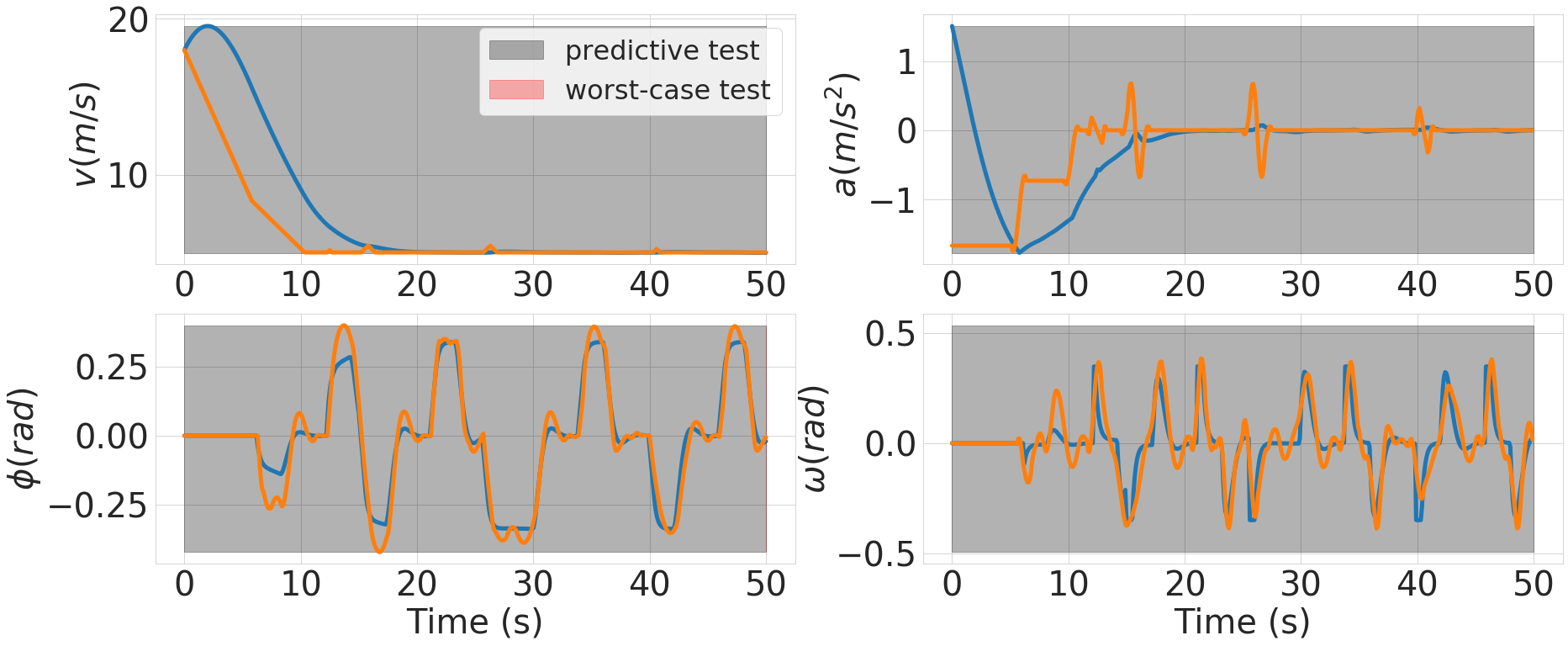}
  \caption{Vehicle velocity (top left), heading angle (top bottom), acceleration control (top right), and steering angle control (bottom right).}
  \label{fig:cib_annoying2}
\end{subfigure}
\caption{An adversarial POV persistently blocks an SV follower with a conservative parameterization.}
\label{fig:cib_annoying}
\end{figure}

\begin{figure}[h]
\centering
\begin{subfigure}{.4\textwidth}
  \centering
  \includegraphics[width=1\textwidth]{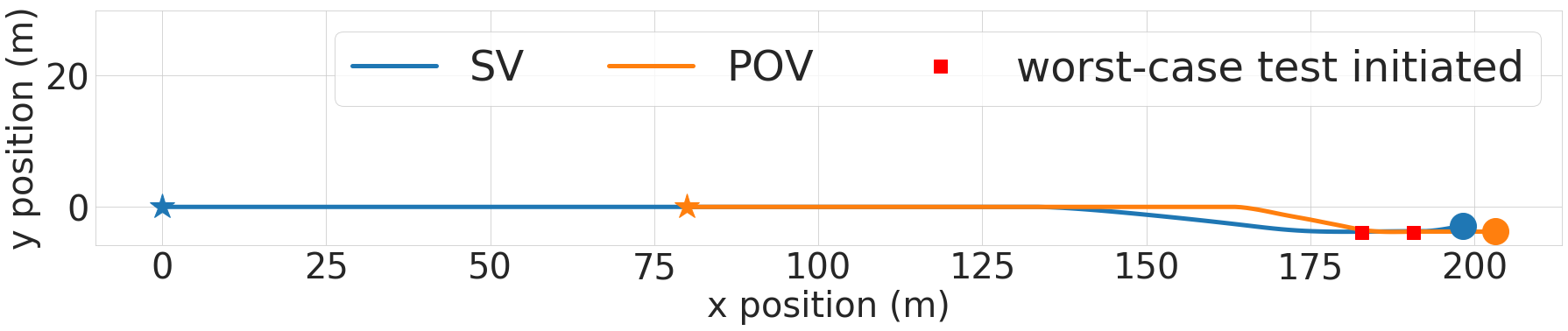}
  \caption{Vehicle position trajectories.}
  \label{fig:cib_crash_smallc1}
\end{subfigure}
\begin{subfigure}{.4\textwidth}
  \centering
  \includegraphics[width=1\textwidth]{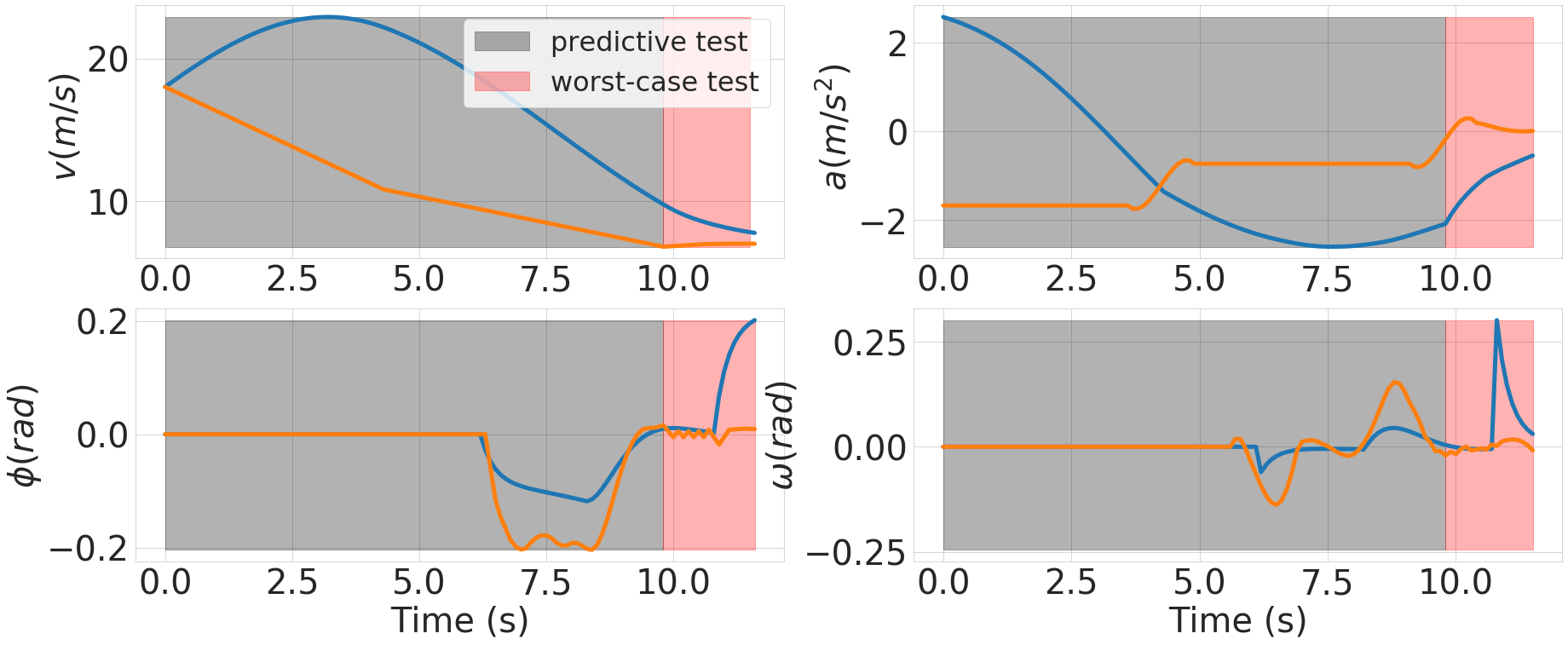}
  \caption{Vehicle velocity (top left), heading angle (top bottom), acceleration control (top right), and steering angle control (bottom right).}
  \label{fig:cib_crash_smallc2}
\end{subfigure}
\caption{An adversarial POV crashes the SV follower with the capture diameter set to $c=7$.}
\label{fig:cib_crash_smallc}
\vspace{-3mm}
\end{figure}

\begin{figure}[t]
\centering
\vspace{3mm}
\begin{subfigure}{.4\textwidth}
  \centering
  \includegraphics[width=1\textwidth]{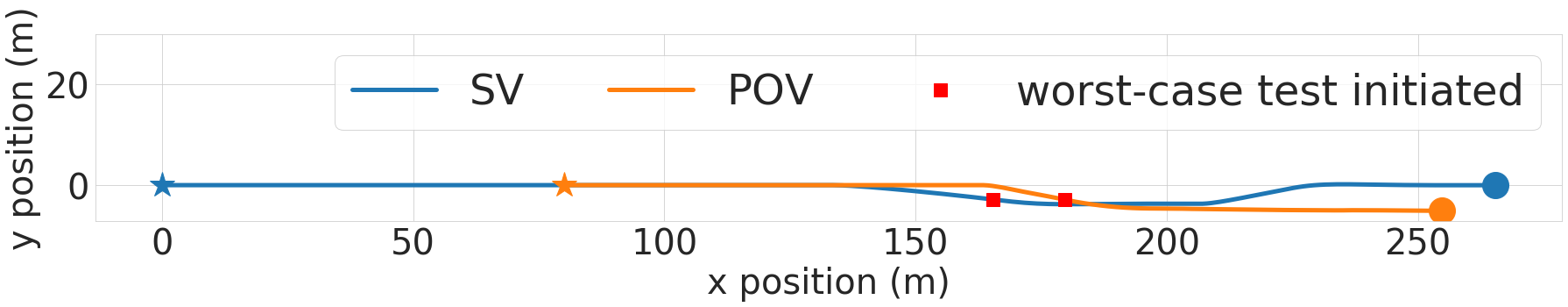}
  \caption{Vehicle position trajectories.}
  \label{fig:cib_nocrash_bigc1}
\end{subfigure}
\begin{subfigure}{.4\textwidth}
  \centering
  \includegraphics[width=1\textwidth]{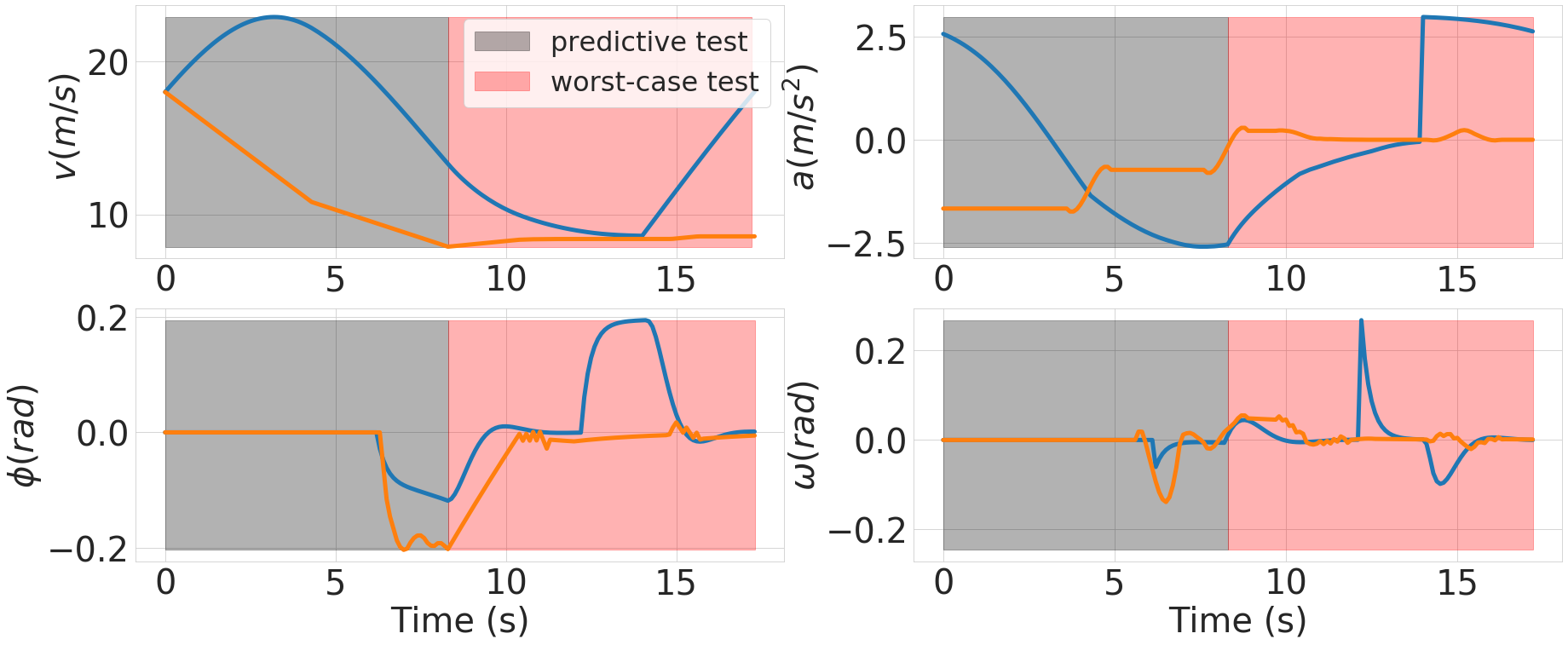}
  \caption{Vehicle velocity (top left), heading angle (top bottom), acceleration control (top right), and steering angle control (bottom right).}
  \label{fig:cib_nocrash_bigc2}
\end{subfigure}
\caption{An adversarial POV fails to force the SV follower into a crash with the capture diameter set to $c=12$.}
\label{fig:cib_nocrash_bigc}
\vspace{-3mm}
\end{figure}

\begin{figure}[h!]
\centering
\vspace{3mm}
\begin{subfigure}{.4\textwidth}
  \centering
  \includegraphics[width=1\textwidth]{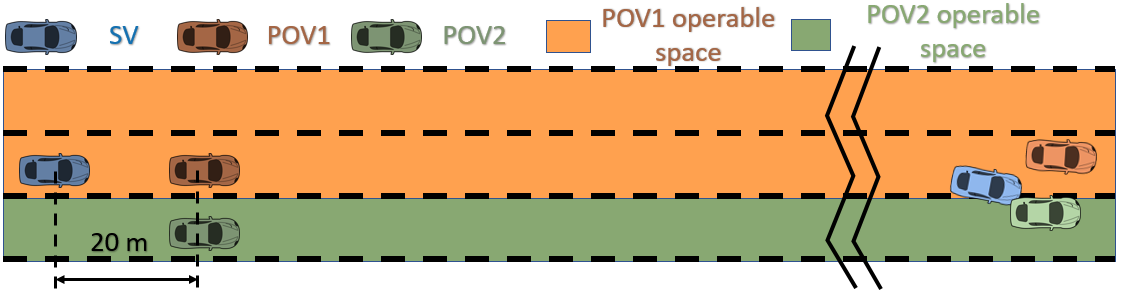}
  \caption{A conceptional plot of the scenario configuration.}
  \label{fig:multi_veh_concept_1}
\end{subfigure}
\begin{subfigure}{.4\textwidth}
  \centering
  \includegraphics[width=1\textwidth]{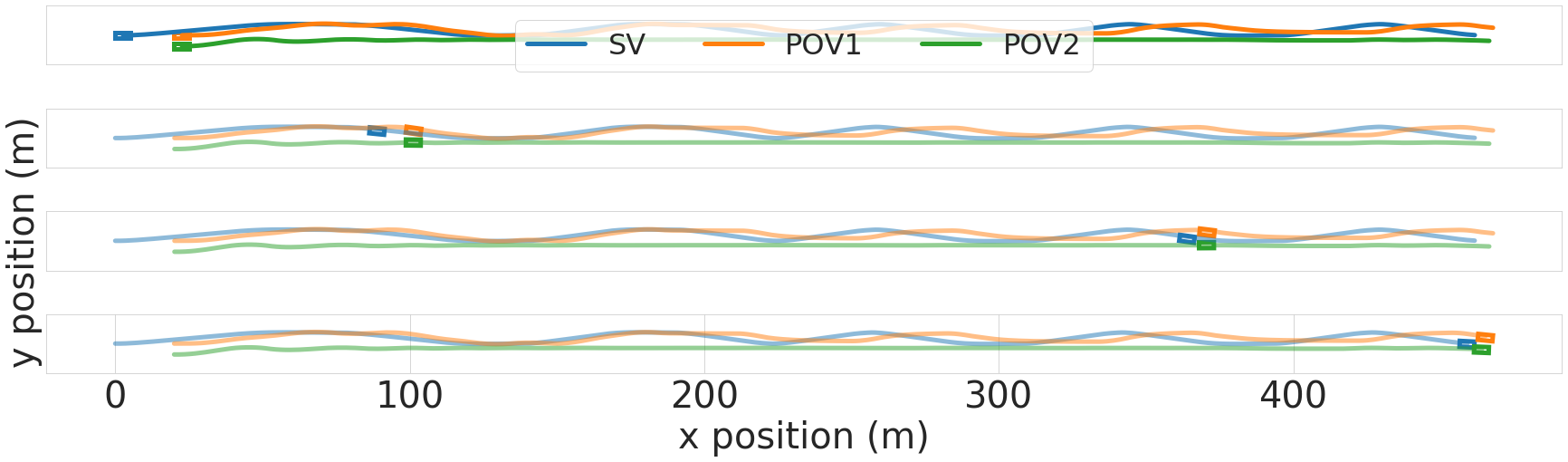}
  \caption{Vehicle position trajectories.}
  \label{fig:multi_apov1}
\end{subfigure}
\begin{subfigure}{.4\textwidth}
  \centering
  \includegraphics[width=1\textwidth]{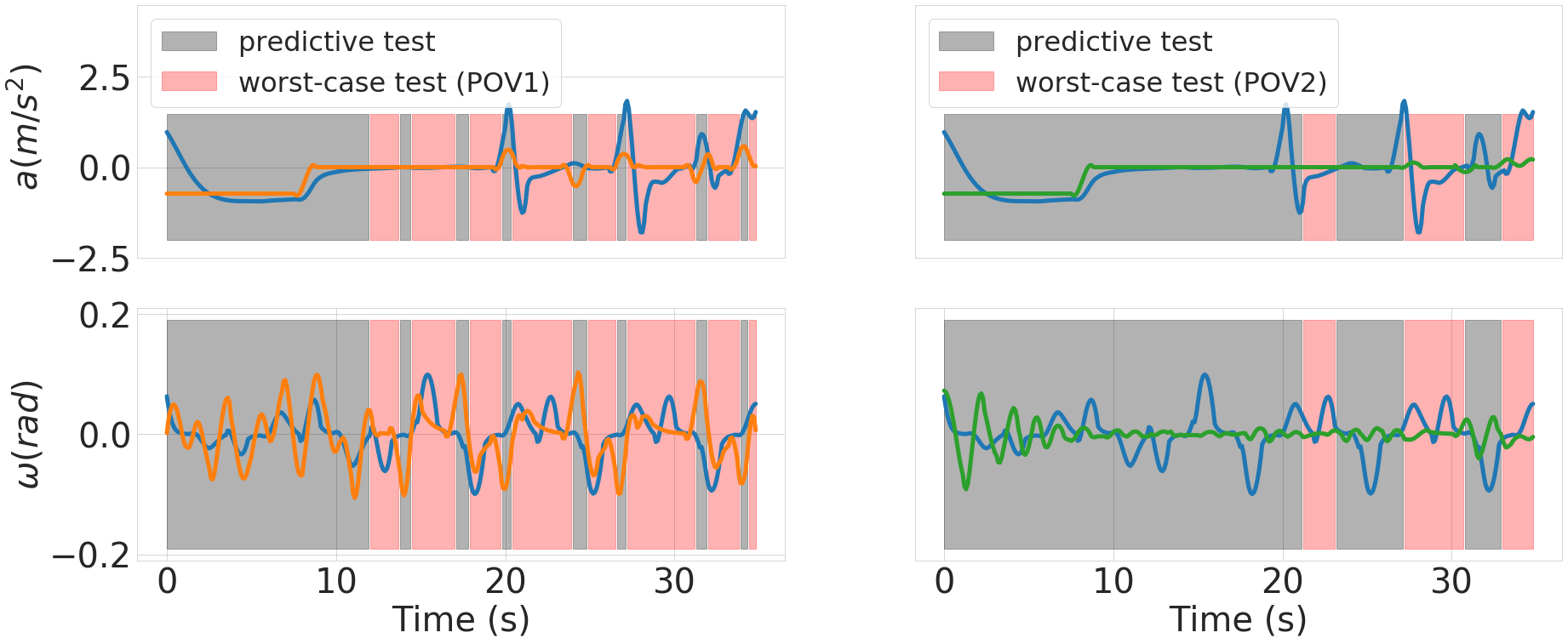}
  \caption{Acceleration control (top row), and steering angle control (bottom row) of the SV and each adversarial POV.}
  \label{fig:multi_apov2}
\end{subfigure}
\caption{Two cooperative lead-POVs interact with the SV follower: SV ends up with a rear-end collision against POV2 due to the aggressive choice of lead gap acceptance during the single lane-change stage.}
\label{fig:multi_apov}
\vspace{-5mm}
\end{figure}

Throughout the pairwise interactive scenarios presented in this section, we consider the same initialization condition $\mathbf{\sigma}_0$ inspired by the standard Crash Imminent Brake (CIB) system test as shown in Fig.~\ref{fig:cib_concept}. Both SV and POV are confined to the three-lane operable domain, with extra state constraints for restricting the POV to initiate a rear-end collision against the SV within the same lane. The admissible velocity range is confined to $v \in [5, 45] (m/s)$. Vehicles are assumed identical with a length of $5 m$ and a width of $2 m$. Lane width is set to $3.7 m$. The adversarial planning algorithm and all vehicle motion controllers are executed at $10$ Hz. A scenario terminates at $50$ seconds after the initialization, or earlier if a vehicle-to-vehicle collision is detected. Unless stated otherwise, the mentioned configurations remain the same for all other examples. Simulation results with various self-driving policies of different hyper-parameters are shown in Fig.~\ref{fig:cib_annoying}, Fig.~\ref{fig:cib_crash_smallc}, and Fig.~\ref{fig:cib_nocrash_bigc}. Here it is worth emphasizing the performance comparison between Fig.~\ref{fig:cib_crash_smallc} and Fig.~\ref{fig:cib_nocrash_bigc}. The SV policy, initialization conditions and state-action constraints are all set identically in both tests with the only difference lying in the choice of the capture diameter $c$. In Fig.~\ref{fig:cib_crash_smallc} we use a smaller value of $c=7$.  The POV switches from the predictive planning mode to worst-case planning mode about $10$ seconds after the test initialization and the collision occurs $2.1$ seconds later. On the other hand, in Fig.~\ref{fig:cib_nocrash_bigc} we take a larger value of $c=12$, which makes it easier to initiate the worst-case planning mode ($2$ seconds earlier than the case in Fig.~\ref{fig:cib_crash_smallc}). However, the relatively large capture space makes the scenario run less severe than the example in Fig.~\ref{fig:cib_crash_smallc} and fails to force a vehicle-to-vehicle collision. Furthermore, note that the SV performs an abrupt acceleration of $2.6 m/s^2$ at $t=14.2s$ during a single lane-change stage to its left. This behavior is significantly outside the expectation as the worst-case planning algorithm assumes the maximum longitudinal acceleration of the SV is only $0.67 m/s^2$. As we have mentioned in Section~\ref{sec:method}, although such an extreme evasive maneuver is deemed aggressive by common sense, it leads the SV to a lower-risk driving status in terms of collision avoidance. Hence, the proposed online adversarial framework succeeds in creating a sufficiently dangerous scenario such that the proper response of the SV can be tested.
\subsubsection{Multi-POV scenario}

\begin{figure}[h]
\centering
\begin{subfigure}{.4\textwidth}
  \centering
  \includegraphics[width=1\textwidth]{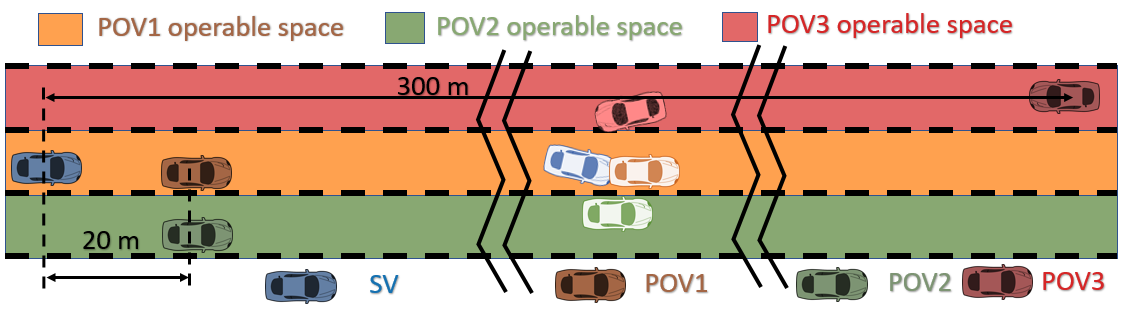}
  \caption{A conceptional plot of the scenario configuration.}
  \label{fig:multi_veh_concept_3}
\end{subfigure}
\begin{subfigure}{.4\textwidth}
  \centering
  \includegraphics[width=1\textwidth]{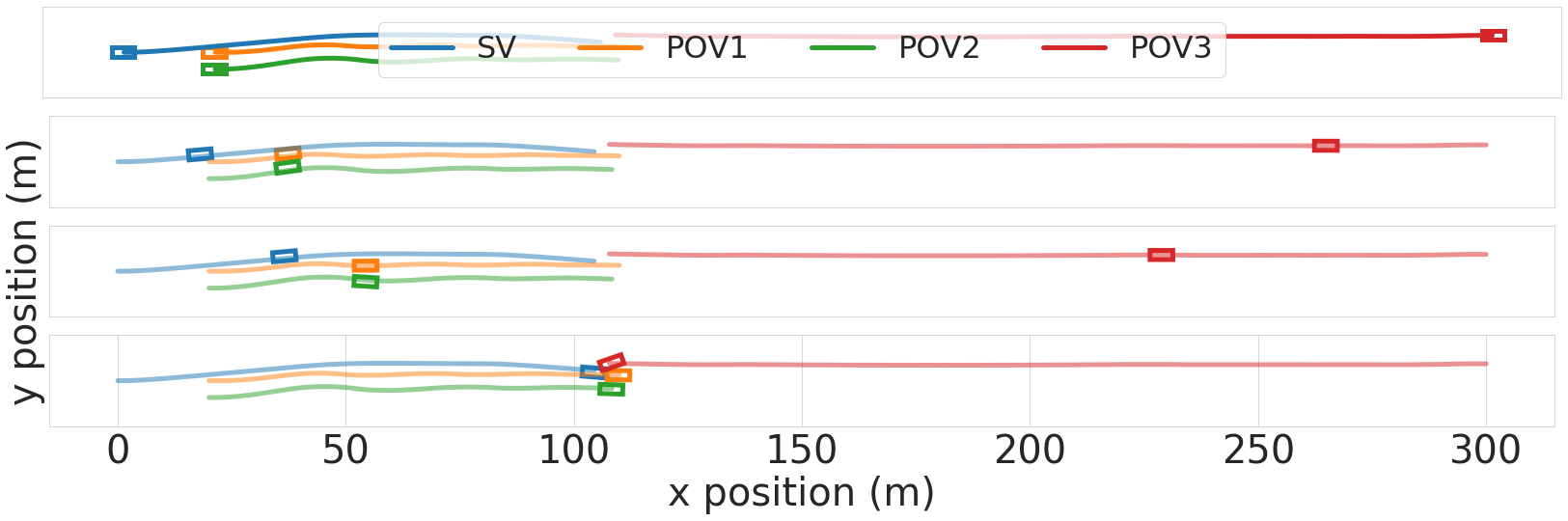}
  \caption{Vehicle position trajectories.}
  \label{fig:multi_apov31}
\end{subfigure}
\begin{subfigure}{.4\textwidth}
  \centering
  \includegraphics[width=1\textwidth]{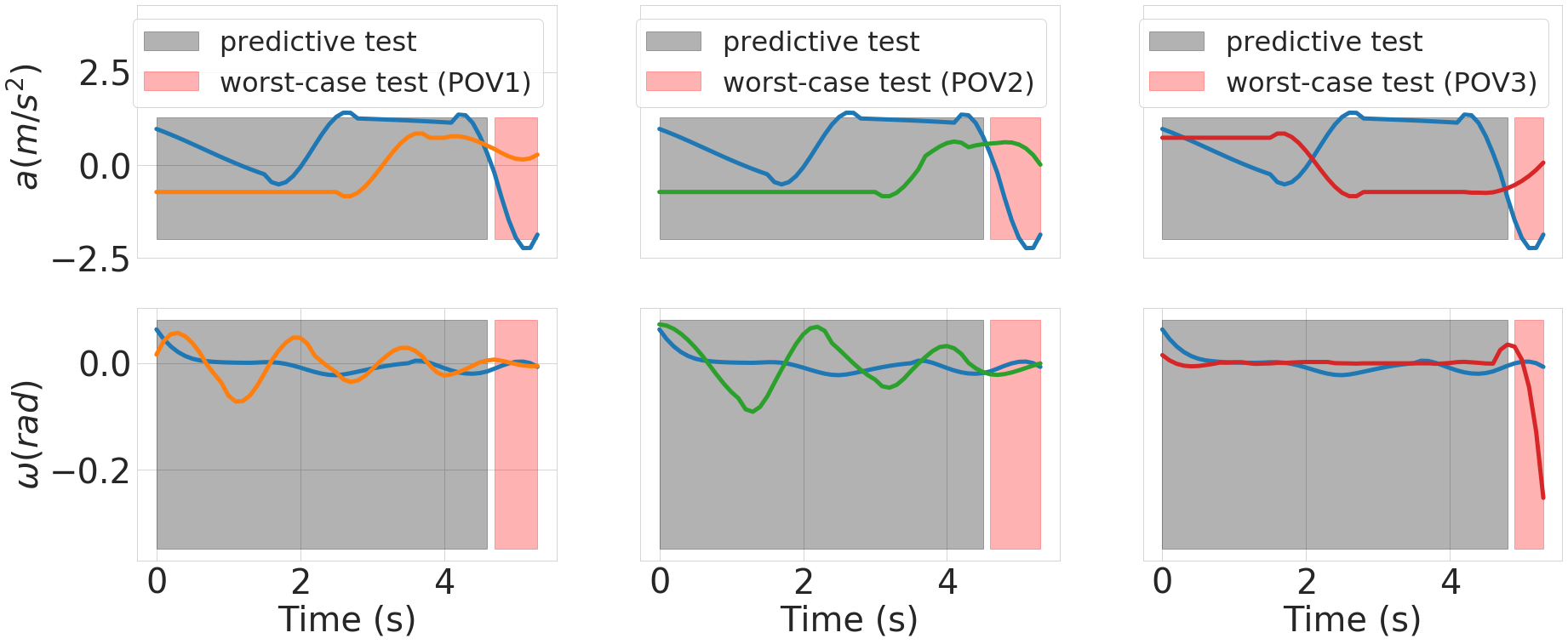}
  \caption{Acceleration control (top row), and steering angle control (bottom row) of SV and each adversarial POV.}
  \label{fig:multi_apov32}
\end{subfigure}
\caption{Three cooperative POV's (two lead-POV's and one on-coming POV) interact with the SV follower: SV performs a single lane-change to avoid collision against the on-comping POV3, ends up with a rear-end collision against POV1.}
\label{fig:multi_apov3}
\end{figure}

\begin{figure}[h]
\centering
\begin{subfigure}{.4\textwidth}
  \centering
  \includegraphics[width=1\textwidth]{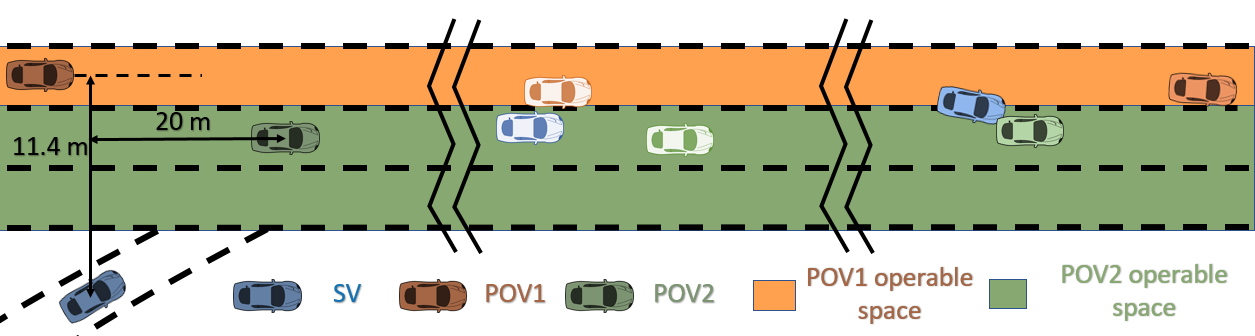}
  \caption{A conceptional plot of the scenario configuration}
  \label{fig:multi_veh_concept_2}
\end{subfigure}
\begin{subfigure}{.4\textwidth}
  \centering
  \includegraphics[width=1\textwidth]{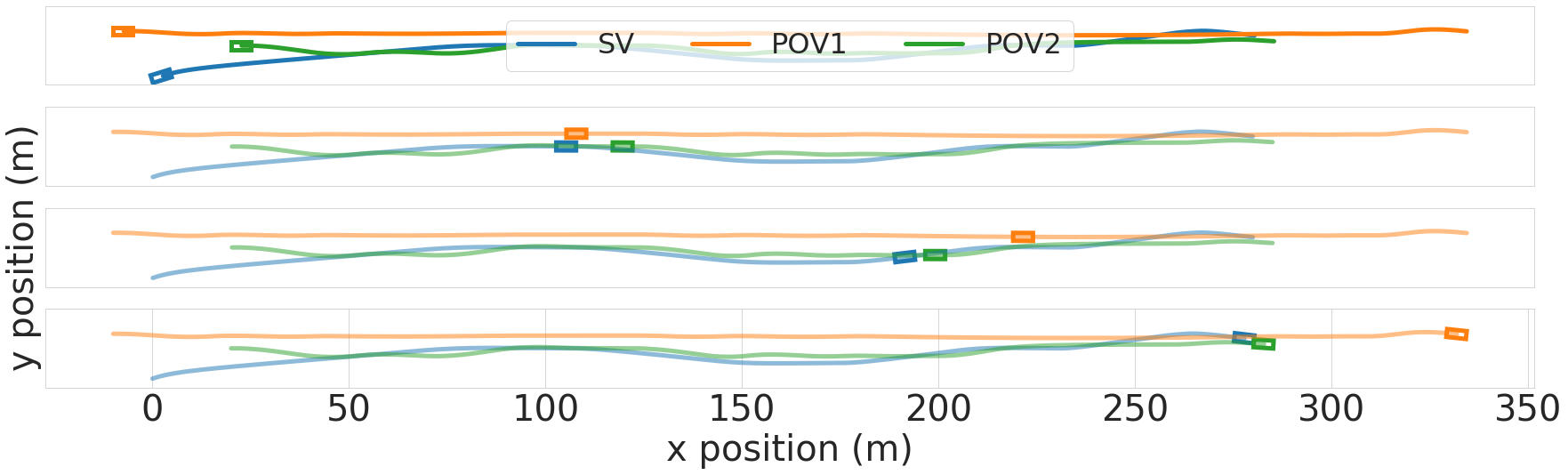}
  \caption{Vehicle position trajectories.}
  \label{fig:multi_apovRAMP1}
\end{subfigure}
\begin{subfigure}{.4\textwidth}
  \centering
  \includegraphics[width=1\textwidth]{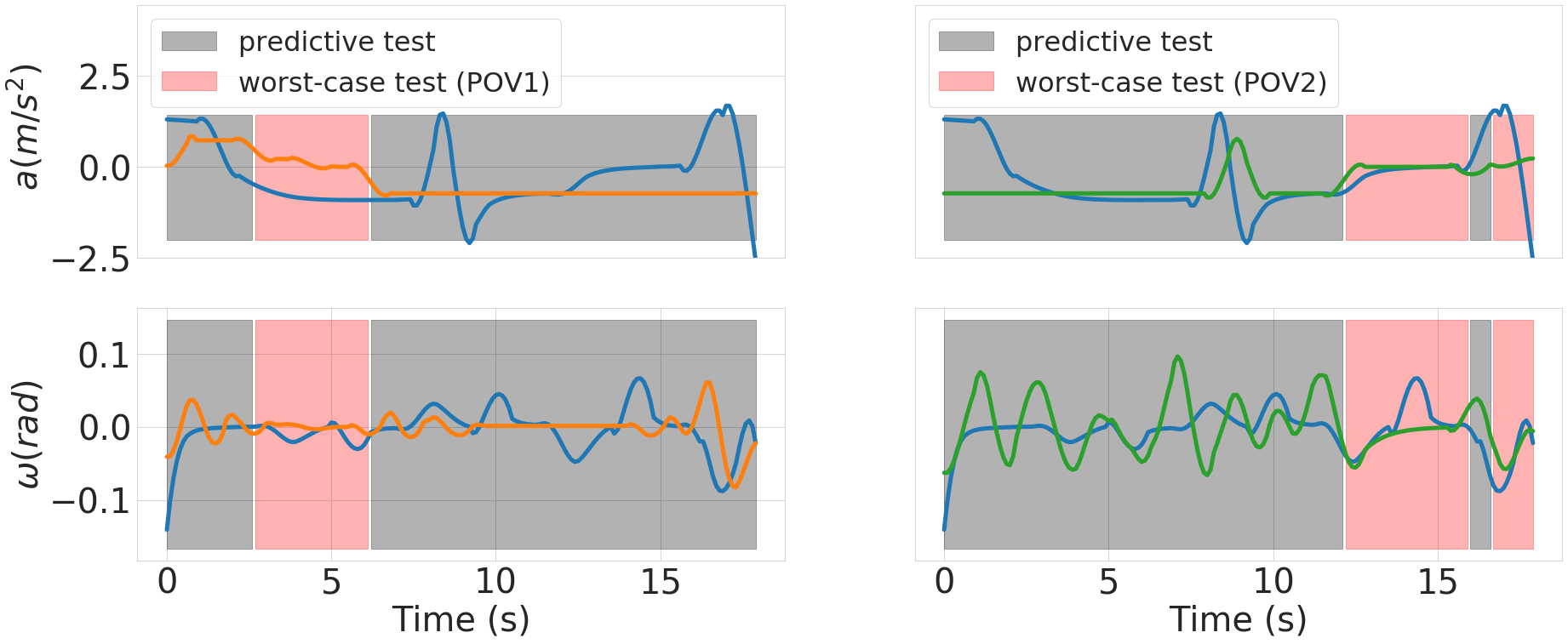}
  \caption{Acceleration control (top row), and steering angle control (bottom row) of SV and each adversarial POV.}
  \label{fig:multi_apovRAMP2}
\end{subfigure}
\caption{Two cooperative POV's interact with the SV merging from a side ramp: SV ends up with a rear-end collision against POV2.}
\label{fig:multi_apovRAMP}
\end{figure}

We further present three multi-POV scenarios as shown in Fig.~\ref{fig:multi_apov}, Fig.~\ref{fig:multi_apovRAMP}, and Fig.~\ref{fig:multi_apov3}. Throughout all the examples, each POV is assigned with a unique operable space that does not overlap with other POV's. The capture diameter is set to $c=7$ for all examples. Admissible velocity range is modified to $v \in [12,45] (m/s)$. The assumed admissible action space for SV and all POV's are identical during the adversarial trajectory planning stage, but in practice, the SV is often more capable than POV both longitudinally and laterally. This is particularly obvious in Fig.~\ref{fig:multi_apov} where the worst-case planning transits back to the predictive planning stage 7 times for POV1 before the SV colliding to POV2 from the rear side. Intuitively, one can refer to Fig.~\ref{fig:multi_apov2} where the SV is performing active single lane-change and double lane-change maneuvers before the collision occurs. Fig.~\ref{fig:multi_apovRAMP} and Fig.~\ref{fig:multi_apov3} give two other examples of how adversarial POV's can cooperate to enforce safety-critical scenarios.


\subsection{Testing human drivers}
\begin{figure}[t!]
\centering
\vspace{3mm}
  \includegraphics[width=0.75\linewidth]{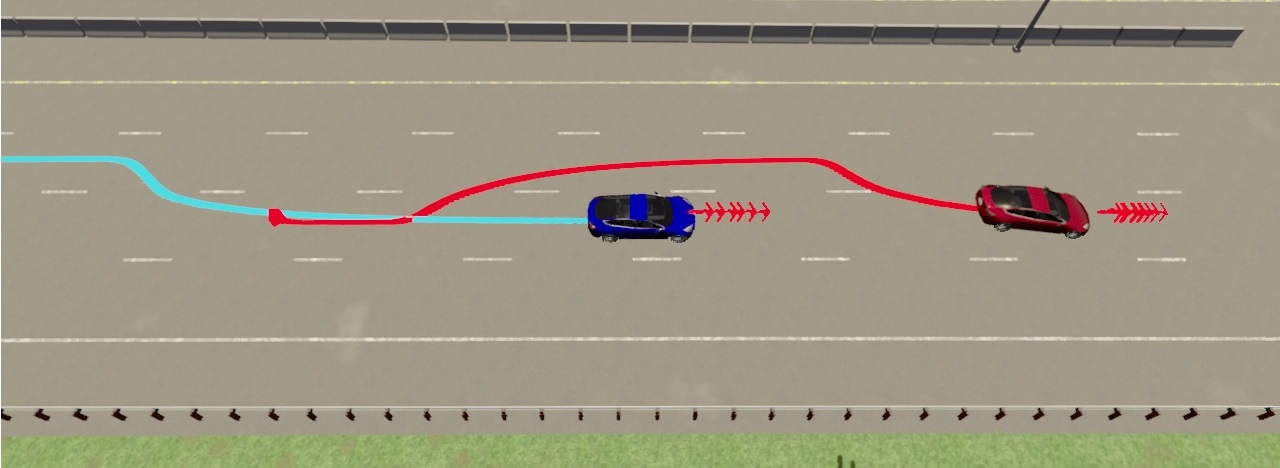}
  \caption{Top view of a pairwise interactive scenario in CARLA, the POV is red and the SV is blue.}
\label{fig:carla}
\vspace{-4mm}
\end{figure}

The proposed method is also implemented in the CARLA simulator. We replicate several of the initial test runs in the previous subsection with similar parameters. Fig.~\ref{fig:carla} shows a top view in the simulator from the initial condition shown in Fig.~\ref{fig:cib_concept}, where the trajectory followed by both the SV (blue) and the POV (red) is shown on-screen. The waypoints from both vehicles for a time horizon of 2 seconds are also drawn.
Additionally, this implementation allows a human driver to take over the SV control while the POV is dynamically reacting to the SV's actions. It is shown that the predictive planning has allowed the POV to react appropriately to provoke near-collision situations even when the SV's policy is completely unexpected (human driver).

\section{Conclusions} \label{sec:concl}
This paper presents an online adversarial framework for the scenario-based testing of operational vehicle safety. The proposed method generates sufficiently dangerous testing scenarios in an efficient and controlled manner. It is applicable for the safety evaluation of human-driven, as well as ADS and ADAS equipped, vehicles. Various simulated examples are also presented to show the empirical effectiveness. It is of future interest to extend the method to different environment configurations such as intersections, roundabout, and parking-lot. We will also improve the methodology with real-world experiments in proving ground tests.

\section*{Acknowledgment}
This work was funded by the United States Department of Transportation under award number 69A3551747111 for Mobility21: the National University Transportation Center for Improving Mobility.  Any findings, conclusions, or recommendations expressed herein are those of the authors and do not necessarily reflect the views of the United States Department of Transportation, Carnegie Mellon University, or The Ohio State University.


\bibliographystyle{IEEEtran}
\bibliography{IEEEabrv,bib}

\begin{thebibliography}{10}
\providecommand{\url}[1]{#1}
\csname url@samestyle\endcsname
\providecommand{\newblock}{\relax}
\providecommand{\bibinfo}[2]{#2}
\providecommand{\BIBentrySTDinterwordspacing}{\spaceskip=0pt\relax}
\providecommand{\BIBentryALTinterwordstretchfactor}{4}
\providecommand{\BIBentryALTinterwordspacing}{\spaceskip=\fontdimen2\font plus
\BIBentryALTinterwordstretchfactor\fontdimen3\font minus
  \fontdimen4\font\relax}
\providecommand{\BIBforeignlanguage}[2]{{%
\expandafter\ifx\csname l@#1\endcsname\relax
\typeout{** WARNING: IEEEtran.bst: No hyphenation pattern has been}%
\typeout{** loaded for the language `#1'. Using the pattern for}%
\typeout{** the default language instead.}%
\else
\language=\csname l@#1\endcsname
\fi
#2}}
\providecommand{\BIBdecl}{\relax}
\BIBdecl

\bibitem{paul2016advanced}
A.~Paul, R.~Chauhan, R.~Srivastava, and M.~Baruah, ``Advanced driver assistance
  systems,'' SAE Technical Paper, Tech. Rep., 2016.

\bibitem{thorn2018framework}
E.~Thorn, S.~C. Kimmel, M.~Chaka, B.~A. Hamilton \emph{et~al.}, ``A framework
  for automated driving system testable cases and scenarios,'' United States.
  Department of Transportation. National Highway Traffic Safety, Tech. Rep.,
  2018.

\bibitem{feng2020testing}
S.~Feng, Y.~Feng, C.~Yu, Y.~Zhang, and H.~X. Liu, ``Testing scenario library
  generation for connected and automated vehicles, part i: Methodology,''
  \emph{IEEE Transactions on Intelligent Transportation Systems}, 2020.

\bibitem{nhtsa2019tja}
``Traffic jam assist ({TJA}) system confirmation test,'' DOT HS 812 759, Tech.
  Rep., 2019.

\bibitem{nhtsa2010cib}
C.~A.~M. Partnership, ``Crash imminent braking ({CIB}) first annual report,''
  DOT HS 811 340, Tech. Rep., 2010.

\bibitem{klischat2019generating}
M.~Klischat and M.~Althoff, ``Generating critical test scenarios for automated
  vehicles with evolutionary algorithms,'' in \emph{2019 IEEE Intelligent
  Vehicles Symposium (IV)}.\hskip 1em plus 0.5em minus 0.4em\relax IEEE, 2019,
  pp. 2352--2358.

\bibitem{chen2020adversarial}
B.~Chen and L.~Li, ``Adversarial evaluation of autonomous vehicles in
  lane-change scenarios,'' \emph{arXiv preprint arXiv:2004.06531}, 2020.

\bibitem{ding2020learning}
W.~Ding, M.~Xu, and D.~Zhao, ``Learning to collide: An adaptive safety-critical
  scenarios generating method,'' \emph{arXiv preprint arXiv:2003.01197}, 2020.

\bibitem{camacho2013model}
E.~F. Camacho and C.~B. Alba, \emph{Model predictive control}.\hskip 1em plus
  0.5em minus 0.4em\relax Springer Science \& Business Media, 2013.

\bibitem{ho1965differential}
Y.~Ho, A.~Bryson, and S.~Baron, ``Differential games and optimal
  pursuit-evasion strategies,'' \emph{IEEE Transactions on Automatic Control},
  vol.~10, no.~4, pp. 385--389, 1965.

\bibitem{mitchell2007comparing}
I.~M. Mitchell, ``Comparing forward and backward reachability as tools for
  safety analysis,'' in \emph{International Workshop on Hybrid Systems:
  Computation and Control}.\hskip 1em plus 0.5em minus 0.4em\relax Springer,
  2007, pp. 428--443.

\bibitem{wensing2017template}
P.~Wensing and S.~Revzen, ``Template models for control,'' \emph{Bioinspired
  Legged Locomotion. Elsevier}, pp. 240--266, 2017.

\bibitem{weng2020model}
B.~Weng, S.~J. Rao, E.~Deosthale, S.~Schnelle, and F.~Barickman, ``{Model
  Predictive Instantaneous Safety Metric for Evaluation of Automated Driving
  Systems},'' in \emph{Proceedings of the 31st IEEE Intelligent Vehicles
  Symposium (IV'20)}, June 2020.

\bibitem{wieber2006trajectory}
P.-B. Wieber, ``Trajectory free linear model predictive control for stable
  walking in the presence of strong perturbations,'' in \emph{2006 6th IEEE-RAS
  International Conference on Humanoid Robots}.\hskip 1em plus 0.5em minus
  0.4em\relax IEEE, 2006, pp. 137--142.

\bibitem{rajamani2011vehicle}
R.~Rajamani, \emph{Vehicle dynamics and control}.\hskip 1em plus 0.5em minus
  0.4em\relax Springer Science \& Business Media, 2011.

\bibitem{junietz2018criticality}
P.~Junietz, F.~Bonakdar, B.~Klamann, and H.~Winner, ``Criticality metric for
  the safety validation of automated driving using model predictive trajectory
  optimization,'' in \emph{2018 21st International Conference on Intelligent
  Transportation Systems (ITSC)}.\hskip 1em plus 0.5em minus 0.4em\relax IEEE,
  2018, pp. 60--65.

\bibitem{erlien2015shared}
S.~M. Erlien, ``Shared vehicle control using safe driving envelopes for
  obstacle avoidance and stability,'' Ph.D. dissertation, Stanford University,
  2015.

\bibitem{Bokanowski2010}
O.~Bokanowski, N.~Forcadel, and H.~Zidani, ``{Reachability and minimal times
  for state constrained nonlinear problems without any controllability
  assumption},'' \emph{SIAM Journal on Control and Optimization}, vol.~48,
  no.~7, pp. 4292--4316, 2010.

\bibitem{houenou2013vehicle}
A.~Houenou, P.~Bonnifait, V.~Cherfaoui, and W.~Yao, ``Vehicle trajectory
  prediction based on motion model and maneuver recognition,'' in \emph{2013
  IEEE/RSJ international conference on intelligent robots and systems}.\hskip
  1em plus 0.5em minus 0.4em\relax IEEE, 2013, pp. 4363--4369.

\bibitem{koschi2017spot}
M.~Koschi and M.~Althoff, ``{SPOT}: A tool for set-based prediction of traffic
  participants,'' in \emph{2017 IEEE Intelligent Vehicles Symposium
  (IV)}.\hskip 1em plus 0.5em minus 0.4em\relax IEEE, 2017, pp. 1686--1693.

\bibitem{petrovskaya2008model}
A.~Petrovskaya and S.~Thrun, ``Model based vehicle tracking for autonomous
  driving in urban environments,'' \emph{Proceedings of robotics: science and
  systems IV, Zurich, Switzerland}, vol.~34, 2008.

\bibitem{capito2020optical}
L.~Capito, K.~Redmill, and U.~Ozguner, ``Optical flow based visual potential
  field for autonomous driving,'' \emph{Proceedings of the 31st IEEE
  Intelligent Vehicles Symposium (IV'20)}, June 2020.

\bibitem{park2013game}
J.~Park, A.~Kurt, and U.~Ozguner, ``A game theoretic approach to controller
  design for cyber-physical systems: collision avoidance,'' in \emph{2013
  ACM/IEEE International Conference on Cyber-Physical Systems (ICCPS)}.\hskip
  1em plus 0.5em minus 0.4em\relax IEEE Computer Society, 2013, pp. 254--254.

\bibitem{bojarski2016end}
M.~Bojarski, D.~Del~Testa, D.~Dworakowski, B.~Firner, B.~Flepp, P.~Goyal, L.~D.
  Jackel, M.~Monfort, U.~Muller, J.~Zhang \emph{et~al.}, ``End to end learning
  for self-driving cars,'' \emph{arXiv preprint arXiv:1604.07316}, 2016.

\bibitem{yurtsever2020integrating}
E.~Yurtsever, L.~Capito, K.~Redmill, and U.~Ozguner, ``Integrating deep
  reinforcement learning with model-based path planners for automated
  driving,'' \emph{Proceedings of the 31st IEEE Intelligent Vehicles Symposium
  (IV'20)}, June 2020.

\bibitem{lee1976theory}
D.~N. Lee, ``A theory of visual control of braking based on information about
  time-to-collision,'' \emph{Perception}, vol.~5, no.~4, pp. 437--459, 1976.

\bibitem{yang2019examining}
M.~Yang, X.~Wang, and M.~Quddus, ``Examining lane change gap acceptance,
  duration and impact using naturalistic driving data,'' \emph{Transportation
  research part C: emerging technologies}, vol. 104, pp. 317--331, 2019.

\bibitem{swaroop2001review}
D.~Swaroop and K.~Rajagopal, ``A review of constant time headway policy for
  automatic vehicle following,'' in \emph{ITSC 2001. 2001 IEEE Intelligent
  Transportation Systems. Proceedings (Cat. No. 01TH8585)}.\hskip 1em plus
  0.5em minus 0.4em\relax IEEE, 2001, pp. 65--69.

\bibitem{dosovitskiy2017carla}
A.~Dosovitskiy, G.~Ros, F.~Codevilla, A.~Lopez, and V.~Koltun, ``{CARLA}: An
  open urban driving simulator,'' \emph{arXiv preprint arXiv:1711.03938}, 2017.

\bibitem{liebner2012driver}
M.~Liebner, M.~Baumann, F.~Klanner, and C.~Stiller, ``Driver intent inference
  at urban intersections using the intelligent driver model,'' in \emph{2012
  IEEE Intelligent Vehicles Symposium}.\hskip 1em plus 0.5em minus 0.4em\relax
  IEEE, 2012, pp. 1162--1167.

\end{thebibliography}
\end{document}